\documentclass[letterpaper, 10 pt, conference]{ieeeconf}
\IEEEoverridecommandlockouts      
\usepackage[utf8]{inputenc}
\usepackage{cite}
\usepackage{graphicx}
\usepackage{amsmath}
\usepackage{amsfonts}
\usepackage{subcaption}
\usepackage{hyperref}
\usepackage{cite}
\usepackage{caption}
\title{\LARGE Benchmarking Visual-Inertial Odometry in Subterranean Environments Under Sensor Degradation, Miscalibration, and Dynamic Occlusion}
\author{
Yueying Zhu$^{1,\dagger}$,
Xiang Li$^{2,\dagger}$,
Thien-Minh Nguyen$^{3}$,
Xuehe Wang$^{4}$,
Shenghai Yuan$^{1}$
\thanks{ $^{1}$Nanyang Technological University, Singapore.  $^{2}$Dalian University of Technology, China. $^{3}$Queensland University, Australia. $^{4}$Sun Yat-sen University, China. {\small $^{\dagger}$Equal contribution} }%
}

\begin{document}
\maketitle
\thispagestyle{empty}
\pagestyle{empty}

\begin{abstract}
Visual-inertial odometry (VIO) is a core capability for autonomous operation in GPS-denied subterranean environments, yet its reliability can degrade sharply under sensor drift, calibration errors, and dynamic occlusion. Existing evaluations mainly emphasize nominal-condition accuracy, offering limited insight into when practical deployment failures occur. In this work, we present a failure-centric stress-test benchmark for VIO in underground environments using the CERBERUS dataset. We systematically evaluate four representative VIO systems spanning filtering-, optimization-, and learning-based paradigms under nine practical perturbation settings, including IMU bias and noise variation, camera intrinsic and extrinsic drift, and dynamic scene occlusion. Beyond conventional trajectory error, we analyze robustness limits through coverage ratio and failure thresholds, revealing breakdown behaviors that are not captured by nominal-condition performance alone. Our study shows distinct vulnerability patterns across VIO paradigms: some methods are more sensitive to inertial degradation, while others are more affected by geometric miscalibration or dynamic interference. These results provide deployment-oriented guidance for VIO selection, calibration prioritization, and reliable operation in challenging underground scenarios. To support reproducible evaluation and future extensions, we will release the full benchmark scripts and evaluation pipeline.
\end{abstract}

\section{INTRODUCTION}
\bstctlcite{IEEEexample:BSTcontrol}

Visual-Inertial Odometry (VIO) has become a standard technique for autonomous navigation in GPS-denied environments, including extreme scenarios such as underground mining and the DARPA Subterranean Challenge \cite{subt2024,cerberus2022}. By fusing camera measurements with inertial data, VIO estimates a robot's six-degree-of-freedom motion in three-dimensional space \cite{qin2017vins,usenko_visual-inertial_2020,orbslam32021}. While modern systems can be highly accurate in well-lit and richly textured scenes, their field reliability often degrades sharply in harsh real-world conditions \cite{engel2013iccv,qin2019b,stumberg_dm-vio_2022}. Underground environments are particularly challenging due to low illumination, texture-scarce structures, and pervasive dynamic obstacles \cite{degradation,robustbench2018}. These stressors can reduce effective feature support, increase spurious correspondences, and destabilize estimation, culminating in failure outcomes such as tracking loss, map corruption, or divergence.

A central but still poorly quantified question for deployment is: under what conditions does VIO fail? In practice, failures are triggered not only by environmental stressors but also by operational constraints. Sensors are frequently re-mounted for maintenance, exposed to mechanical shocks, or operated with imperfect time synchronization, causing drift in camera intrinsics, camera-IMU extrinsics, and IMU bias/noise parameters \cite{drift2014,fisheye2006}. VIO pipelines often exhibit nonlinear cliff effects: performance may appear stable under small deviations until a threshold is crossed, after which tracking becomes unreliable or collapses. Yet existing evaluations rarely provide actionable failure boundaries, i.e., how much deviation in which parameter (and under which stressor) is sufficient to induce breakdown.

Despite continued advances, prior studies expose several limitations. Many works examine environmental robustness in isolation\cite{liu2024benchmarking,zhangbench2025}, for example, improving performance in low light using event cameras \cite{mueggler2018continuous} or learned features, without quantifying when these methods break under realistic parameter drift \cite{eventvio2006,airslam2025}. Conversely, calibration-oriented studies highlight that focal length drift, distortion mismatch, and IMU bias errors can significantly degrade accuracy, but precise collapse thresholds remain poorly defined and rarely compared across different VIO paradigms \cite{onlinecalib2018,fisheye2006}. Finally, widely used benchmarks such as EuRoC primarily evaluate under comparatively idealized conditions, limiting their ability to inform reliability in extreme deployments \cite{euroc2016}.

To address this gap, we conduct a controlled, failure-centric stress test of four representative VIO systems spanning major paradigms: AirSLAM, OpenVINS, ORB-SLAM3, and SchurVINS \cite{airslam2025,openvins2020,orbslam32021,schurvins2024}. We perturb nine sensor parameters that commonly drift in practice, including four IMU terms (accelerometer and gyroscope biases and noises), three camera intrinsics (focal length, principal point, and distortion coefficient), and two camera-IMU extrinsics (relative rotation and translation) \cite{onlinecalib2018}. To keep the evaluation tractable while revealing interpretable sensitivity trends, we adopt a one-at-a-time (OAT) perturbation protocol rather than a full factorial design. We report failure-aware metrics that remain meaningful under early termination, including a coverage ratio measuring the fraction of the ground-truth trajectory completed before tracking loss \cite{zhang2018traj}, as well as relative RMSE normalized by baseline performance. Beyond outcome metrics, we use observability and numerical-stability diagnostics (e.g., information-matrix rank, Jacobian conditioning, and Hessian definiteness) to connect breakdowns to concrete mechanisms \cite{10003890}. Our results map robustness envelopes and collapse thresholds across paradigms, providing actionable guidance for extreme-environment deployment.
Our contribution can be summarized as:
\begin{itemize}
    \item We establish a stress-test benchmark for visual-inertial odometry in subterranean environments, covering controlled sensor drift and dynamic occlusion on the CERBERUS dataset.
    
    \item We identify robustness limits of four representative VIO systems across nine practical perturbation settings, showing when accurate estimators break under deployment-relevant stress.
    
    \item We provide actionable guidance on VIO selection, calibration tolerance, and operating conditions for underground deployment.
\end{itemize}

\section{Related Work}

Research on SLAM benchmarking in complex environments primarily unfolds across datasets, algorithm evaluation, and framework development.

Research on datasets for underground mine scenarios has largely revolved around the DARPA Subterranean (SubT) Challenge\cite{subt2024}. Beyond the official competition dataset, various research teams have released representative underground environment datasets, including CERBERUS\cite{cerberus2022}, Tunnel-Circuit CTU\cite{CTU1,CTU2}, and Robotic Interestingness\cite{wang2020visual}. These benchmarks effectively compare algorithmic performance under extreme conditions, yet most evaluations lack a core focus on visual inertial odometry (VIO) and lack systematic stress testing. Furthermore, existing mine-scenario benchmarks like OIVIO\cite{OIVIO}, MIN3D\cite{MIN3D1,MIN3D2}, and Chilean\cite{Chilean} exhibit specific limitations, including missing or poorly synchronized IMU data, and visual sequences rendered uninitializable by estimators due to extremely low illumination. 
Mainstream benchmarking efforts have shifted toward developing cloud-based or containerized automated tools\cite{bujanca2019slambench,bujanca2021robust,Bodin2018,Nardi2015}, where SLAMFuse\cite{slamframework2024} introduces the software engineering concept of fuzzing, dynamically injecting brightness/contrast variations or image blur to probe algorithmic boundaries. While this approach identifies visual front-end vulnerabilities, it remains limited to image-level degradation and cannot simulate sensor hardware degradation common in mine tunnel scenarios (e.g., IMU bias drift or internal/external parameter calibration errors). 
Despite existing standard trajectory evaluation tools, current literature lacks a unified framework to couple internal sensor parameter drift with external environmental stressors. Existing tests fail to identify which specific input degradation triggers estimator failure. This paper addresses this gap by conducting sensor noise and dynamic occlusion stress tests on 4 VIO algorithms with CERBERUS dataset.

\section{VIO ALGORITHMS AND DATASETS}
\subsection{Algorithms}

We select four representative VIO algorithms to cover the main design paradigms
(learning-based, filtering, and optimization) and to reflect methods widely
used in real deployments. We evaluate them under identical perturbations to
enable a fair, apples-to-apples robustness comparison:
\begin{itemize}
    \item \textbf{AirSLAM}: Learning-based front-end with strong robustness under challenging visual conditions \cite{airslam2025}.
    \item \textbf{OpenVINS}: Filtering-based VIO with real-time performance and widespread use \cite{openvins2020}.
    \item \textbf{ORB-SLAM3}: Global optimization-based SLAM that jointly estimates poses and IMU states \cite{orbslam32021}.
    \item \textbf{SchurVINS}: Optimization-based VIO using Schur complement for efficient marginalization \cite{schurvins2024}.
\end{itemize}

\subsection{Dataset}
The evaluation is conducted on the CERBERUS dataset, recorded during the DARPA Subterranean Challenge at the Louisville Mega Cavern\cite{cerberus2022}. This dataset represents one of the most demanding benchmarks for mobile robot perception due to its extreme operational conditions. The environment comprises a complex mixture of man-made tunnels and natural cave networks. These settings introduce three simultaneous perceptual stressors: (i) severe non-uniform illumination from onboard LEDs, (ii) repetitive, low-gradient textures on rock surfaces, and (iii) high-frequency mechanical vibrations induced by the quadrupedal gait.

\begin{itemize}
\item \textbf{Robot Platform \& Sensor Suite:} We utilize recordings from four ANYmal C quadrupedal robots. Each platform is equipped with a Sevensense Alphasense Core unit, integrating a factory-calibrated stereo monochrome camera (720×540 px at 20~Hz) and a six-axis IMU (200~Hz). For ground truth generation, a Velodyne VLP16 Puck LITE LiDAR provides high-resolution geometric constraints.
\item \textbf{Dynamic Sequence Extraction:} Unlike conventional pre-clipped benchmarks, the CERBERUS raw recordings contain comprehensive multi-modal streams including LiDAR and secondary sensors. For our VIO stress-testing framework, we specifically extracted the synchronized data from the two monochrome cameras (\textit{cam0} and \textit{cam1}) and the IMU. To ensure evaluation rigor, we manually curated segments of active locomotive motion from the full mission logs, yielding eight high-intensity sequences (\textit{seq0}--\textit{seq7}) totaling approximately 22 minutes. This targeted curation ensures that our metrics reflect true kinematic drift rather than being diluted by stationary intervals.
\item \textbf{Ground-Truth and Trajectories:} CERBERUS adopt the post-processed \textit{CompSLAM post} state estimates as the definitive ground truth reference. This reference is generated by registering VLP16 pointclouds against the DARPA-provided global map. All trajectories are synchronized and exported in the TUM format (timestamp, position, and orientation quaternions), enabling precise absolute trajectory error (ATE) evaluation using the \textit{evo} toolkit.
\end{itemize}
\begin{figure*}[t]
\vspace{8pt}
\centering
\begin{minipage}{0.24\textwidth}
    \centering
    \includegraphics[width=\linewidth]{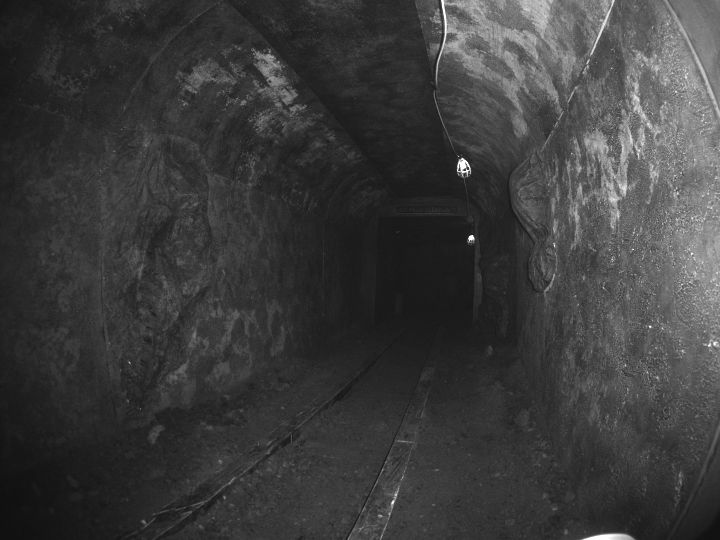}
    \caption*{(a) 0\% (Baseline)}
\end{minipage}
\hfill
\begin{minipage}{0.24\textwidth}
    \centering
    \includegraphics[width=\linewidth]{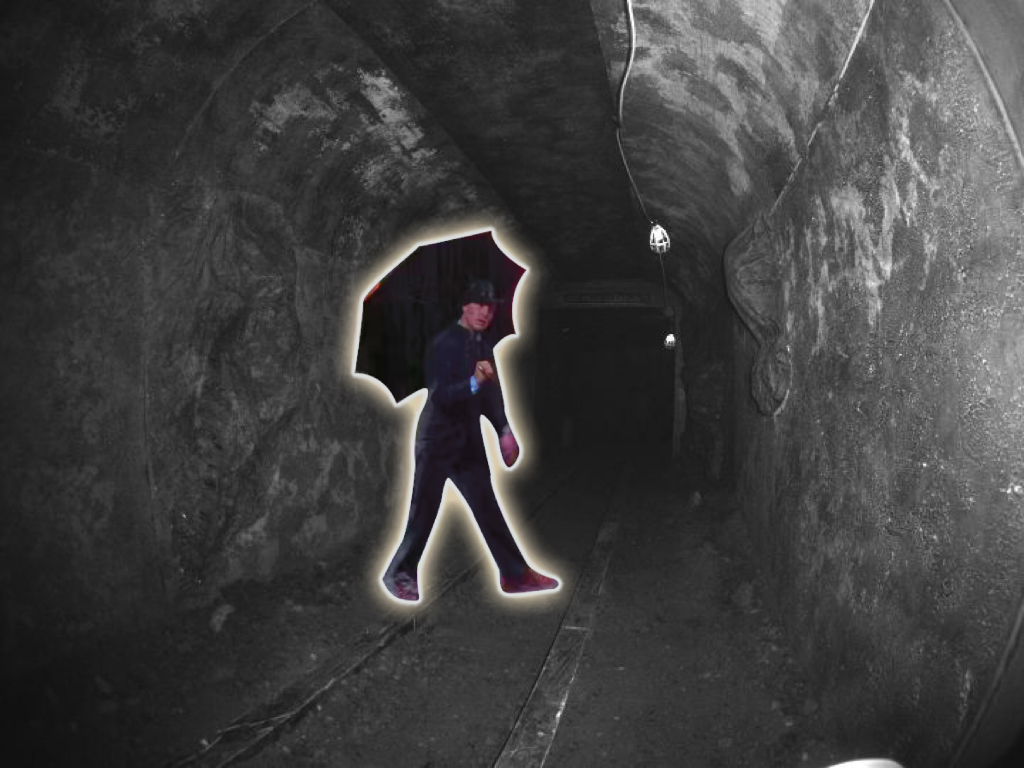}
    \caption*{(b) 3.35\% (Nominal)}
\end{minipage}
\hfill
\begin{minipage}{0.24\textwidth}
    \centering
    \includegraphics[width=\linewidth]{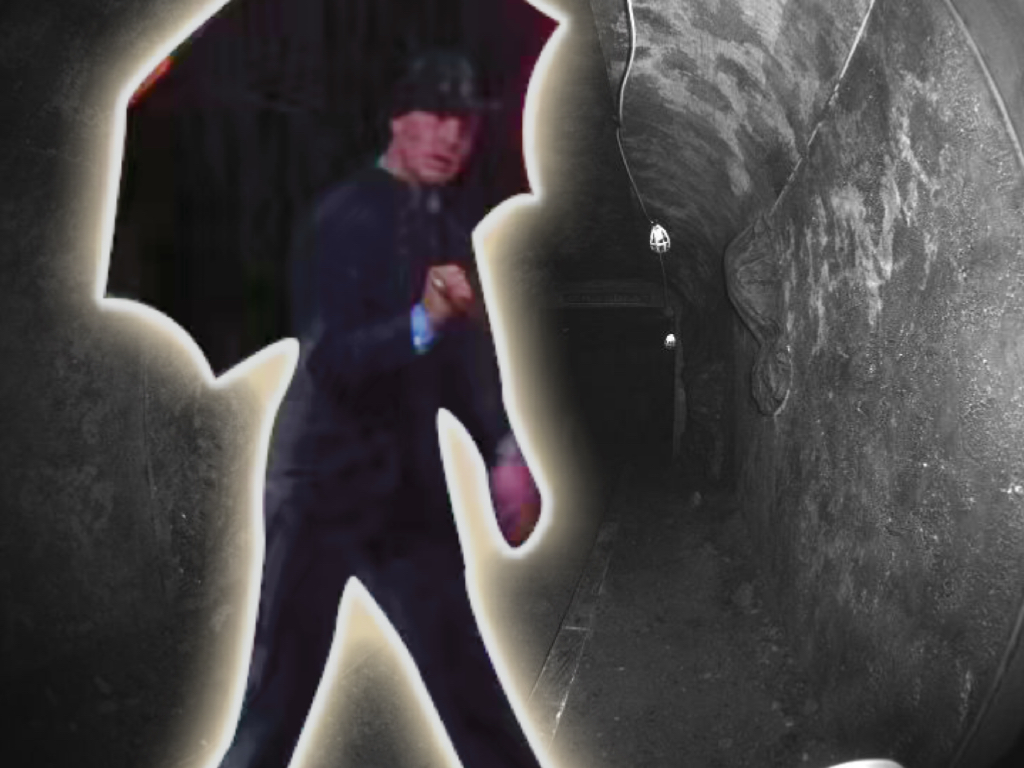}
    \caption*{(c) 30\% (Significant) }
\end{minipage}
\hfill
\begin{minipage}{0.24\textwidth}
    \centering
    \includegraphics[width=\linewidth]{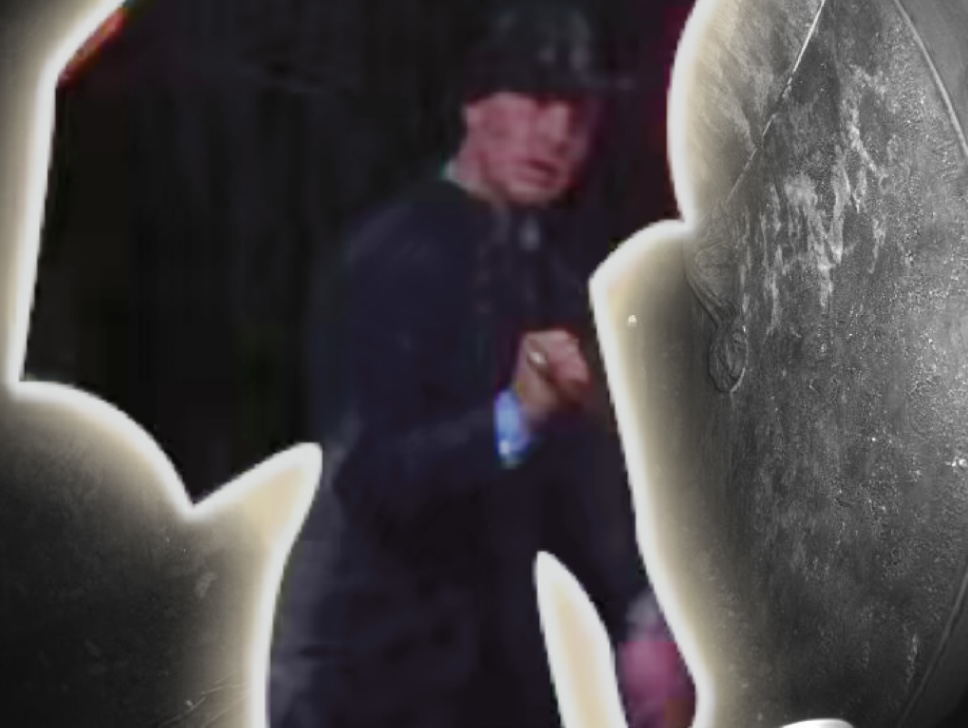}
    \caption*{(d) 60\% (High Occlusion) }
\end{minipage}
\caption{Representative samples of synthetic occlusion masks applied to the CERBERUS subterranean dataset. The masks progress from (a) a baseline clear environment to (b) nominal personnel interference, (c) significant obstruction within the critical poisoning zone, and (d) severe feature starvation. These centered masks simulate the typical visual footprint of personnel and equipment in narrow mining corridors.}
\label{fig:occlusion_samples}
\vspace{-8pt}
\end{figure*}
\section{EXPERIMENTS}

To evaluate the robustness of various VIO/SLAM algorithms during the critical initialization phase, we designed a comprehensive Parameter Stress Analysis. By subjecting the state estimators to controlled perturbations across multiple dimensions using a fixed dataset and ground-truth trajectory, we characterize the sensitivity of each algorithm as a function of perturbation intensity.

\subsection{Sensor Noise and Calibration Stress Tests}

The core of our evaluation is a "one-at-a-time" sensitivity analysis. In each trial, a single parameter dimension is perturbed while all others remain at their nominal values, allowing for the isolation of specific failure modes.The primary metric is the trajectory Root Mean Square Error (RMSE) in meters, calculated after SE(3) alignment with the ground-truth trajectory by EVO. To account for the multi-degree-of-freedom nature of certain parameters (e.g., 3-axis rotation or translation), we employ three random seeds ($s_0, s_1, s_2$). These seeds represent different directional vectors or axes of perturbation, effectively projecting high-dimensional uncertainty into a distance-based scalar analysis for consistent comparison.
To evaluate the system's sensitivity, we sampled 11 perturbation levels for each of the nine critical sensor parameters. These perturbations range from nominal performance to system divergence, as detailed below:

\begin{itemize}
    \item \textbf{Accelerometer Bias ($b_a$):} Scaled by $10^x$, with $x \in \{-2.0, -1.6, -1.2, -0.8, -0.4, 0.0, 0.4, 0.8, 1.2, 1.6, 2.0\}$. Unit: $[m/s^2]$.
    \item \textbf{Gyroscope Bias ($b_g$):} Scaled by the same logarithmic factors $10^x$ as $b_a$. Unit: $[rad/s]$.
    \item \textbf{Accelerometer Noise ($\sigma_a$):} Noise density scaled by $10^x$. Unit: $[m/s^2/\sqrt{Hz}]$.
    \item \textbf{Gyroscope Noise ($\sigma_g$):} Noise density scaled by $10^x$. Unit: $[rad/s/\sqrt{Hz}]$.
    \item \textbf{Extrinsic Rotation ($R_{IC}$):} Rotational deviations $\Delta\theta \in \{0, \pm 0.4, \pm 0.8, \pm 1.2, \pm 1.6, \pm 2.0\}^\circ$ superimposed on the nominal extrinsic matrix along a selected axis.
    \item \textbf{Extrinsic Translation ($t_{IC}$):} Translational shifts $\Delta t \in \{0, \pm 1, \pm 2, \pm 3, \pm 4, \pm 5\}$ cm applied along a specific direction.
    \item \textbf{Focal Length ($f$):} Nominal focal length scaled by $\{0.89, 0.91, 0.93, 0.95, 0.98, 1, 1.02, 1.05, 1.07, 1.1, 1.12\}$.
    \item \textbf{Principal Point ($c_p$):} Pixel-wise offsets $\Delta c \in \{0, \pm 10, \pm 20, \pm 30, \pm 40, \pm 50\}$ px applied to the image center.
    \item \textbf{Radial Distortion ($k_1$):} The distortion coefficient scaled by $\{0.5, 0.58, 0.66, 0.76, 0.87, 1, 1.15, 1.32, 1.51, 1.74, 2\}$.
\end{itemize}

\subsection{Dynamic and Occlusion Resilience Tests}

To systematically evaluate the robustness of VIO algorithms against dynamic interference and limited visibility, we propose a multi-dimensional stress-testing framework. This framework simulates varying degrees of dynamic stress by manipulating two key parameters: the occlusion ratio p and the dynamic factor d.

The dynamic factor d governs the perceived temporal dynamics of the environment, defined by the ratio between the observed frame rate and the original dataset frequency:
\begin{equation}
d = \frac{\mathcal{F}_{mask}}{\mathcal{F}_{dataset}}
\end{equation}
where ${\mathcal{F}_{dataset}} $is the nominal 20~Hz capture rate. By adjusting d, we simulate a wide spectrum of obstacle velocities, ranging from quasi-static motion at low d values to extreme high-speed traversals at the upper bound of the scale.

Simultaneously, centered synthetic masks are applied to the visual stream to simulate the spatial footprint of dynamic obstacles encountered in subterranean operations, as illustrated in Fig. \ref{fig:occlusion_samples}. These occlusion levels are categorized into specific operational scenarios to provide physical context to the numerical results:
\begin{itemize}
\item \textbf{3.35\% (Nominal Personnel):} Represents a single person passing at a distance, which is a standard condition in active mines.
\item \textbf{30\% (Significant Obstruction):} Simulates nearby personnel or equipment, creating a critical ``poisoning zone'' where dynamic features often bypass outlier rejection.
\item \textbf{High Occlusion (over 50\%):} Marks the onset of feature starvation and severe visibility loss, testing the limits of geometric constraints in narrow tunnels.
\end{itemize}

The discrete parameter sets utilized for the sensitivity analysis are summarized as follows:
\begin{itemize}
\item \textbf{Occlusion Rate:} $\{3.35,10,20,30,40,50,60,70,80,90\}\%$.
\item \textbf{Dynamic Factor:} $\{0.05,0.1,0.2,0.5,1,2,5,10,20\}$.
\end{itemize}

\subsection{Evaluation Metrics}

\begin{figure}[t]
\vspace{8pt}
\centering
\includegraphics[width=\columnwidth]{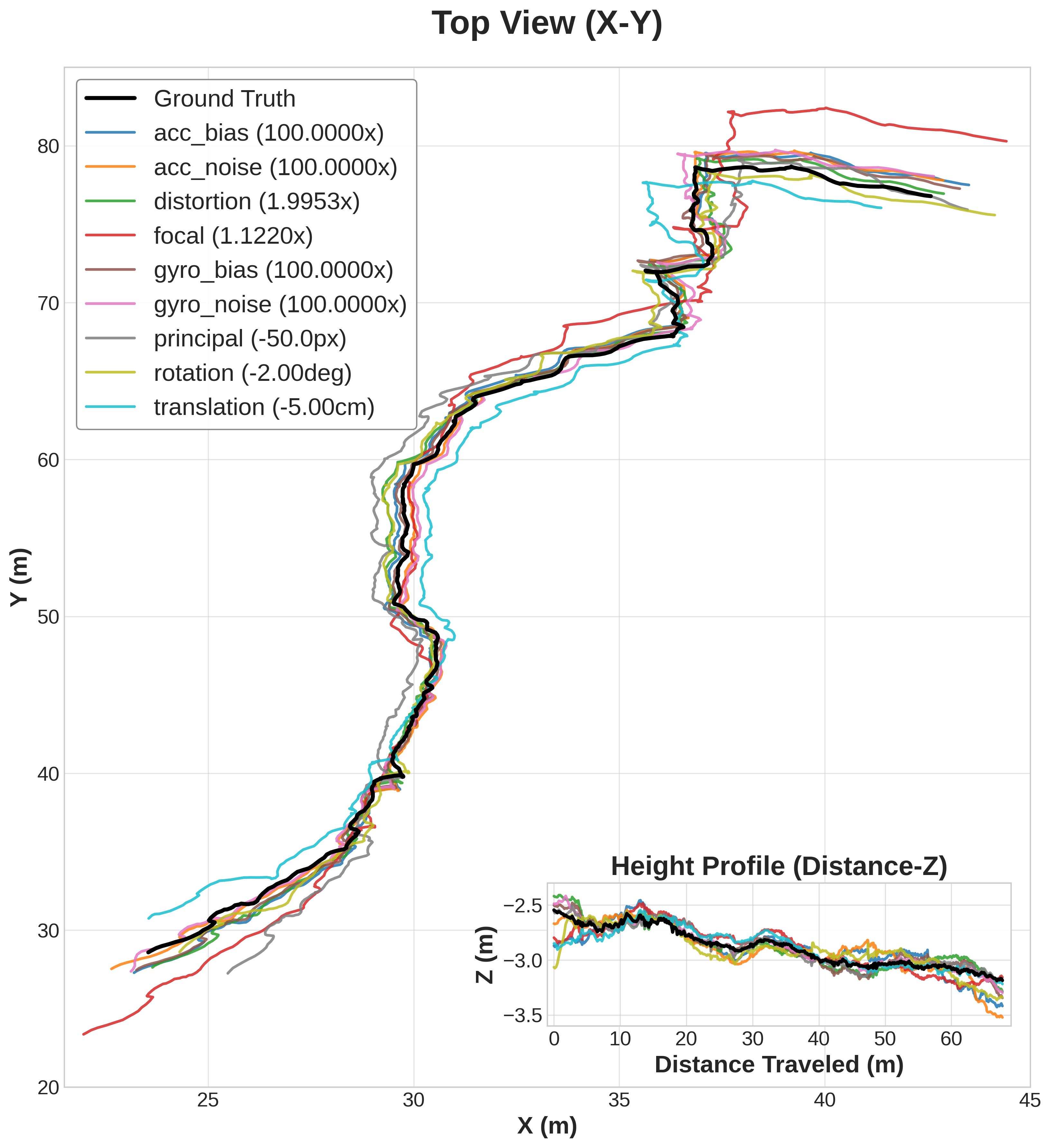}
\caption{Qualitative analysis of trajectory deformations under isolated sensor perturbations. Spatial divergence in the $xy$-plane and vertical susceptibility in the distance-$z$ inset highlight characteristic geometric artifacts, such as scale drift and stochastic oscillations, induced by intrinsic and inertial stressors.}
\label{fig:trajectories_comparison}
\vspace{-8pt}
\end{figure}

Localization fidelity is quantified via the Root Mean Square Error (RMSE) of the $SE(3)$-aligned Absolute Trajectory Error (ATE), which effectively captures the geometric deformations—ranging from scale drift to stochastic oscillations—induced by the sensor perturbations illustrated in Fig. \ref{fig:trajectories_comparison}.

To evaluate mission survivability in subterranean environments, we define the Coverage Ratio $C = d_{\text{track}} / d_{\text{total}}$, where $d_{\text{track}}$ denotes the distance traversed prior to estimator failure, typically precipitated by feature starvation or state divergence.

To mitigate reporting bias from premature termination, a constant drift penalty of 0.025~m/s is integrated into the RMSE for truncated segments, ensuring the metric holistically reflects the entire mission profile.

\begin{figure*}[!t]
\vspace{8pt}
\centering
\includegraphics[width=0.98\textwidth]{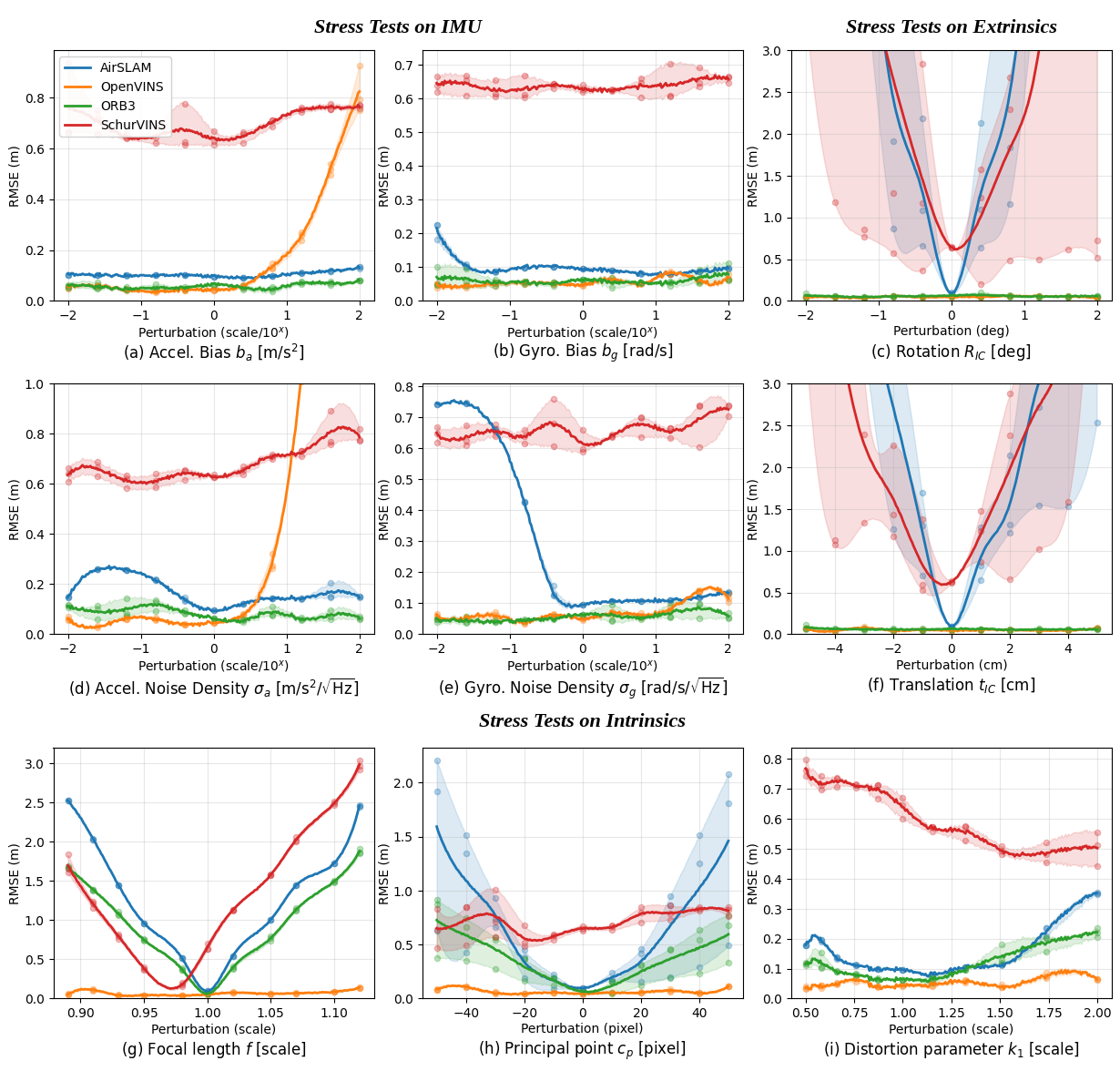}
\caption{Robustness and sensitivity analysis under varying parameter perturbations. We evaluate the positioning accuracy (RMSE) of AirSLAM, OpenVINS, ORB3, and SchurVINS against three categories of system uncertainties: (a-b, d-e) Stress tests on IMU: impact of accelerometer/gyroscope biases ($b_a$, $b_g$) and noise densities ($\sigma_a$, $\sigma_g$) on system stability. (c, f) Stress tests on extrinsics: sensitivity to initial calibration errors in camera-IMU rotation ($R_{IC}$) and translation ($t_{IC}$). (g-i) Stress tests on intrinsics: robustness against inaccuracies in camera focal length ($f$), principal point ($c_p$), and distortion coefficient ($k_1$). Solid lines represent the mean RMSE across multiple trials, while the shaded regions denote the standard deviation. The sharp V-shape curves (e.g., in extrinsics and intrinsics) highlight the system dependency on accurate parameter initialization.}
\label{fig:sensitivity_grid}
\end{figure*}

\begin{figure}[t!]
\vspace{8pt}
\centering
\includegraphics[width=0.98\linewidth]{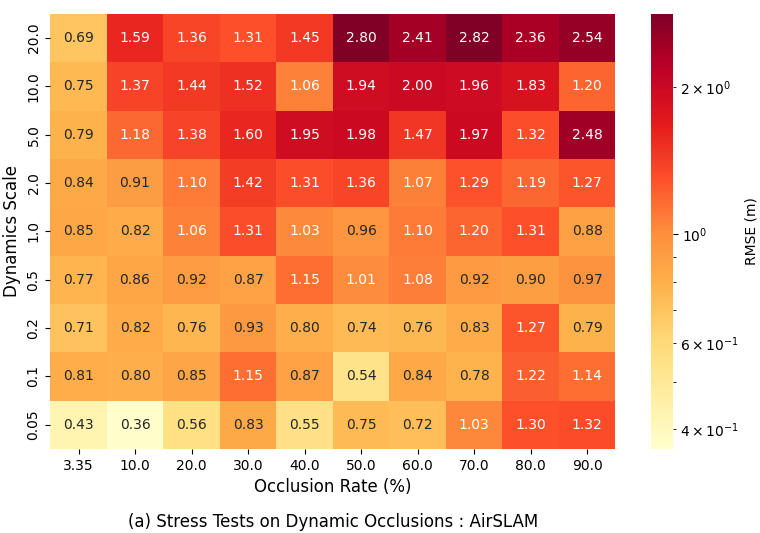} \\
\includegraphics[width=0.98\linewidth]{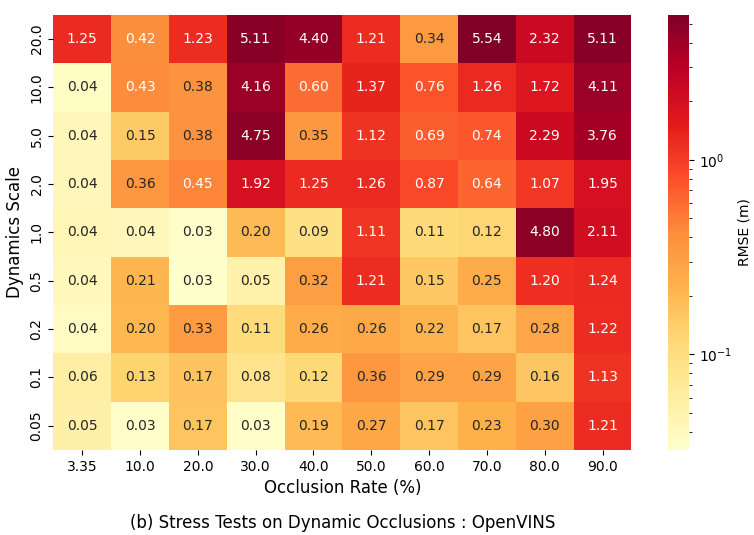} \\
\includegraphics[width=0.98\linewidth]{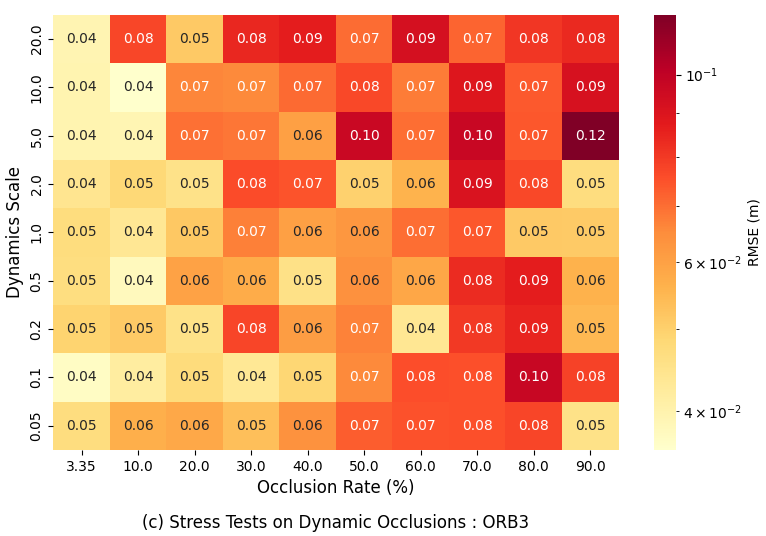}
\caption{Comparative robustness heatmaps for AirSLAM, OpenVINS, and ORB3 evaluated across 270 configurations. The heatmaps illustrate the localization RMSE (meters) as a function of occlusion rates (up to 90.0\%) and dynamics scales ($d$, representing obstacle velocity). A higher $d$ indicates faster movement of personnel within the environment.}
\label{fig:occlusion_dynamics_heatmaps}
\end{figure}

\begin{figure*}[t!]
\vspace{8pt}
\centering
\begin{subfigure}{0.32\textwidth}
    \centering
    \includegraphics[width=\linewidth]{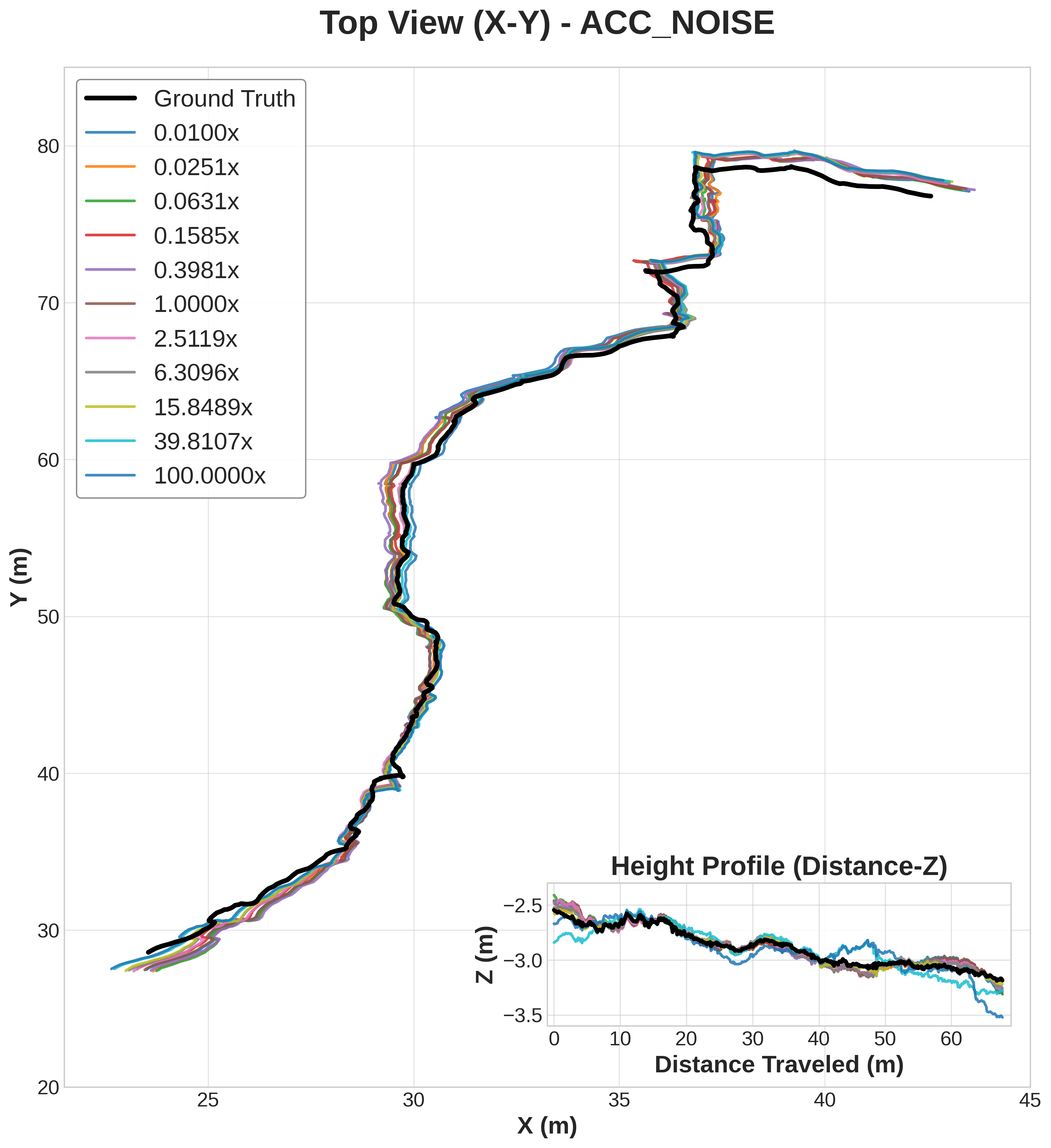}
    \caption{Acc. Noise ($\sigma_a$)}
\end{subfigure}
\hfill
\begin{subfigure}{0.32\textwidth}
    \centering
    \includegraphics[width=\linewidth]{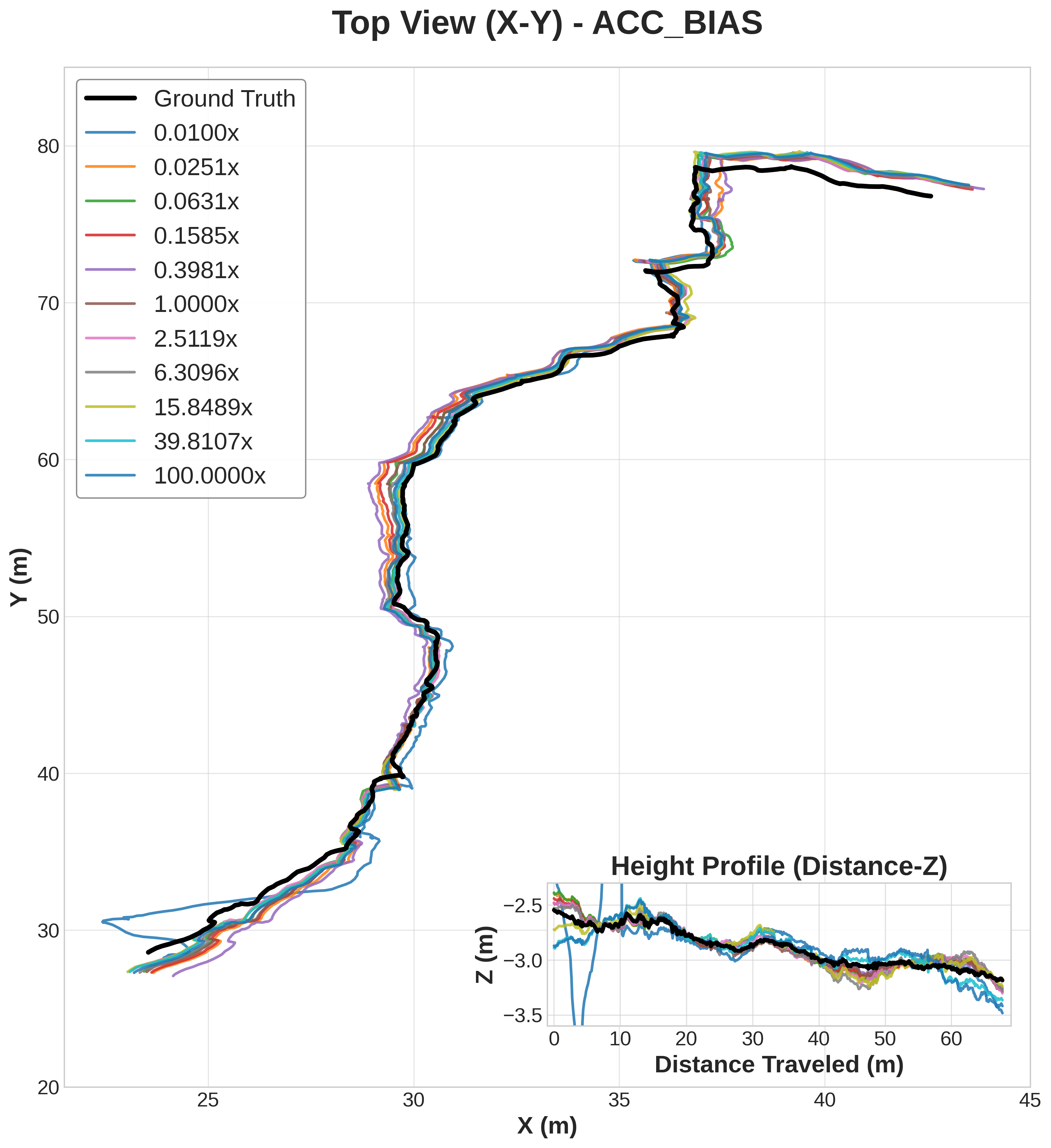}
    \caption{Acc. Bias ($b_a$)}
\end{subfigure}
\hfill
\begin{subfigure}{0.32\textwidth}
    \centering
    \includegraphics[width=\linewidth]{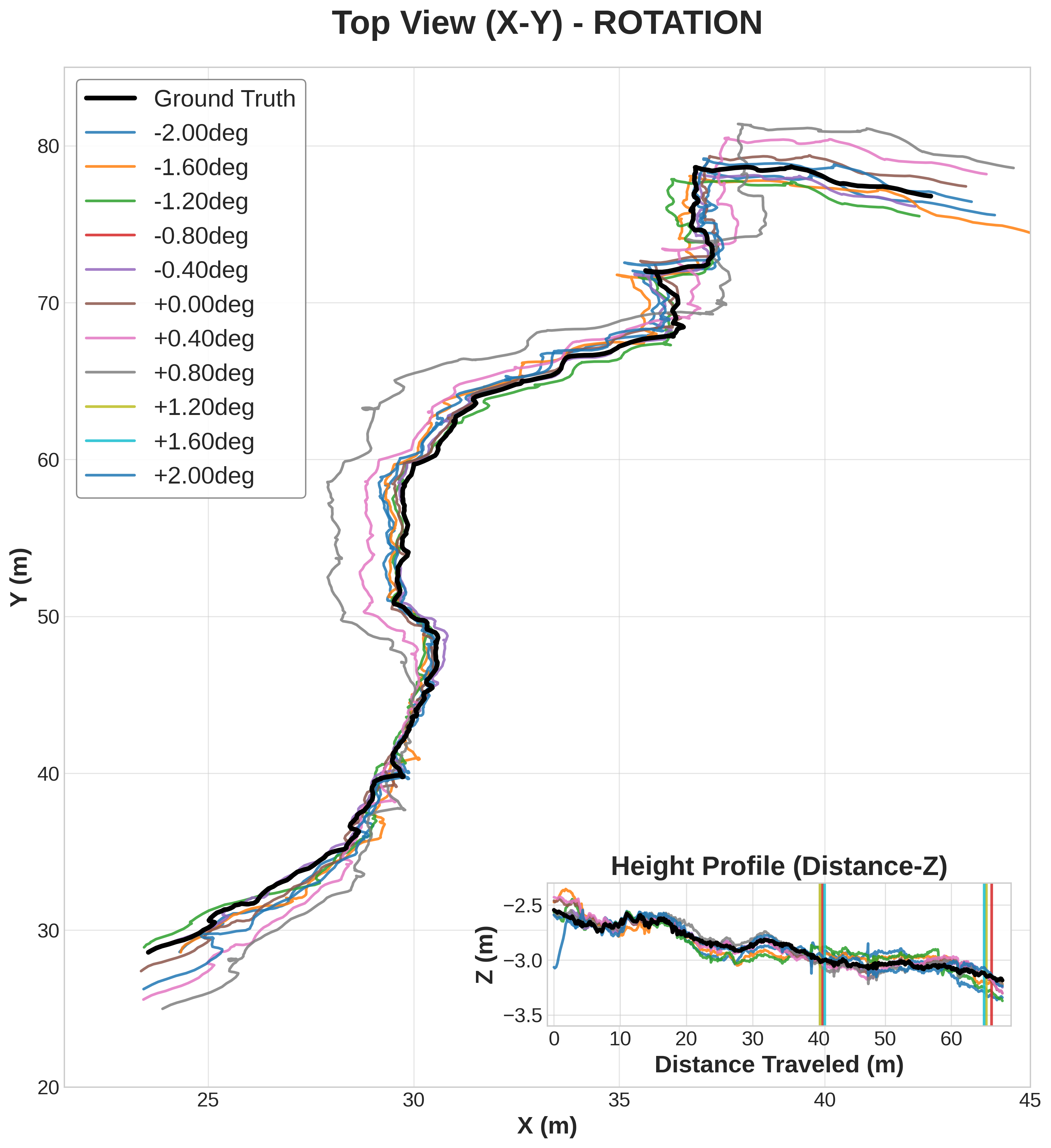}
    \caption{Extrinsic Rot. ($R_{IC}$)}
\end{subfigure}

\begin{subfigure}{0.32\textwidth}
    \centering
    \includegraphics[width=\linewidth]{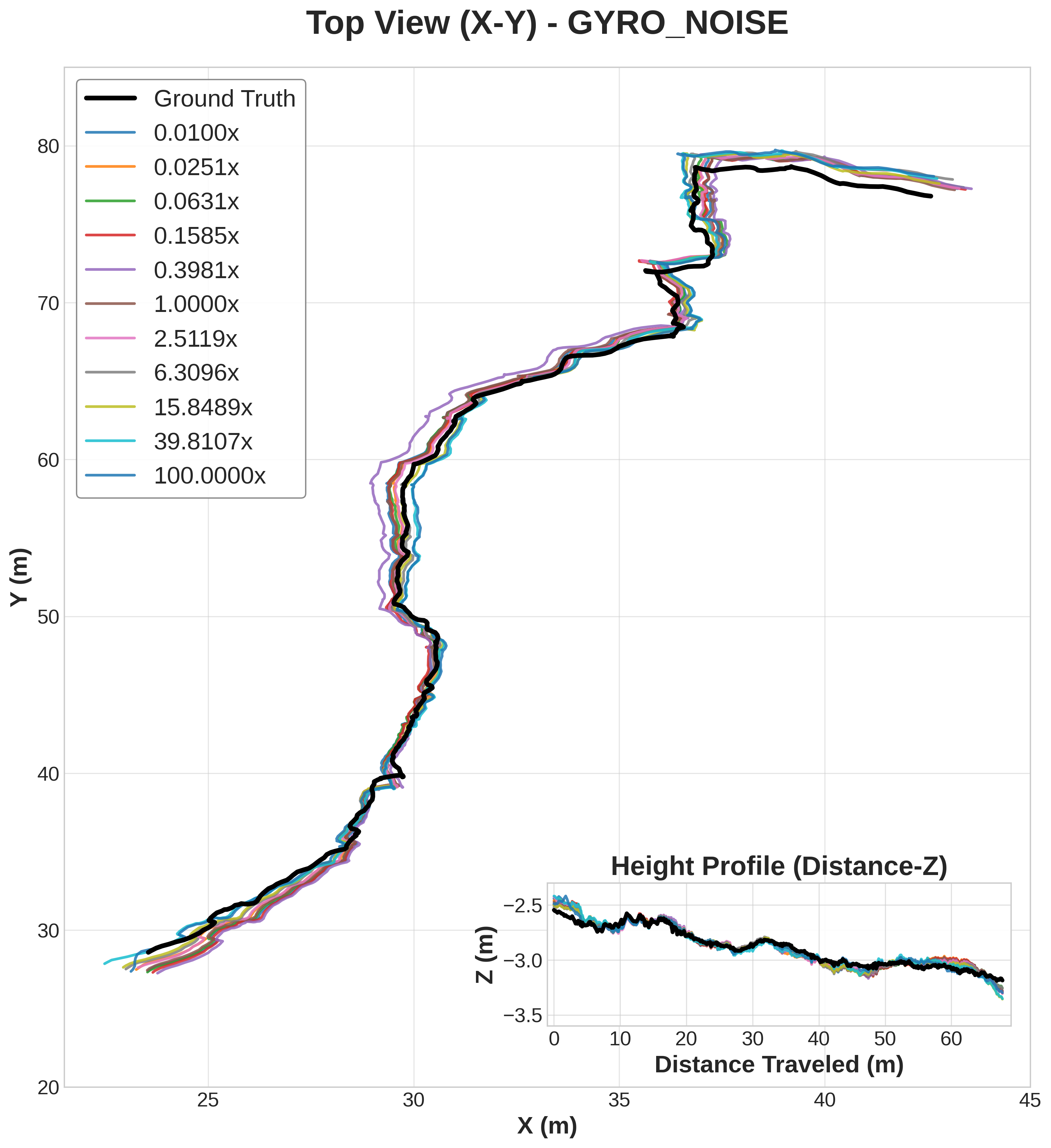}
    \caption{Gyro. Noise ($\sigma_g$)}
\end{subfigure}
\hfill
\begin{subfigure}{0.32\textwidth}
    \centering
    \includegraphics[width=\linewidth]{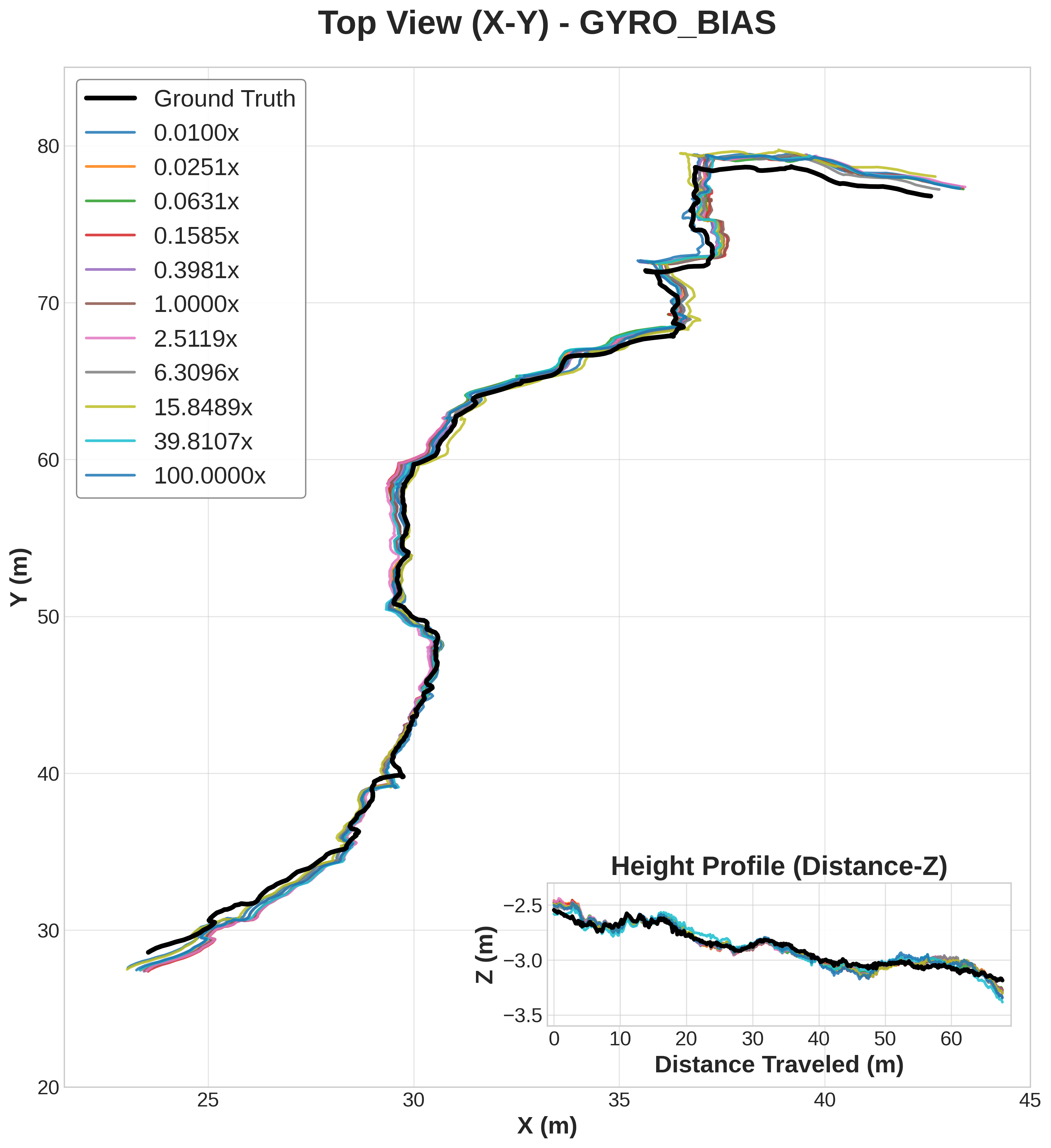}
    \caption{Gyro. Bias ($b_g$)}
\end{subfigure}
\hfill
\begin{subfigure}{0.32\textwidth}
    \centering
    \includegraphics[width=\linewidth]{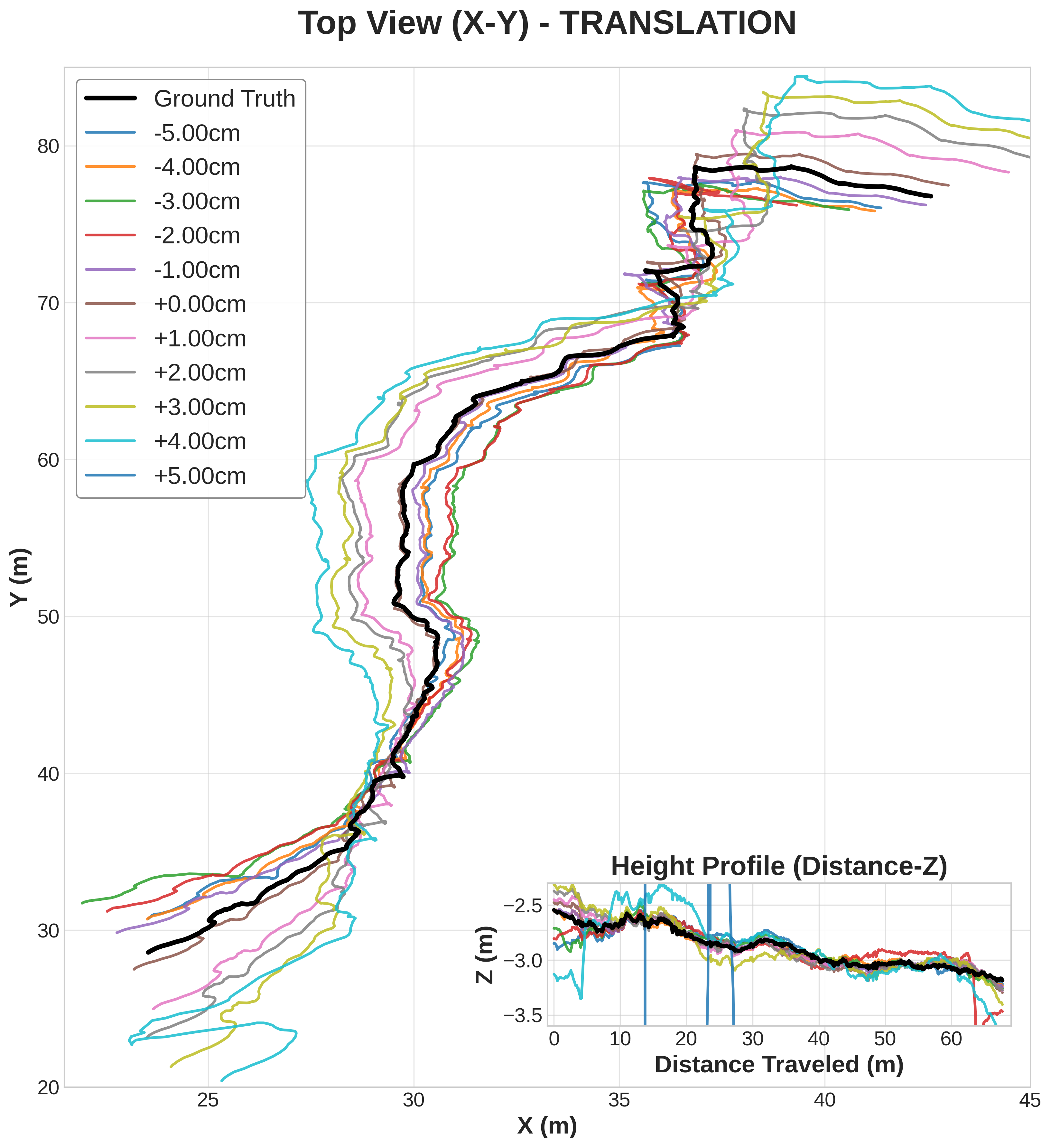}
    \caption{Extrinsic Trans. ($t_{IC}$)}
\end{subfigure}

\begin{subfigure}{0.32\textwidth}
    \centering
    \includegraphics[width=\linewidth]{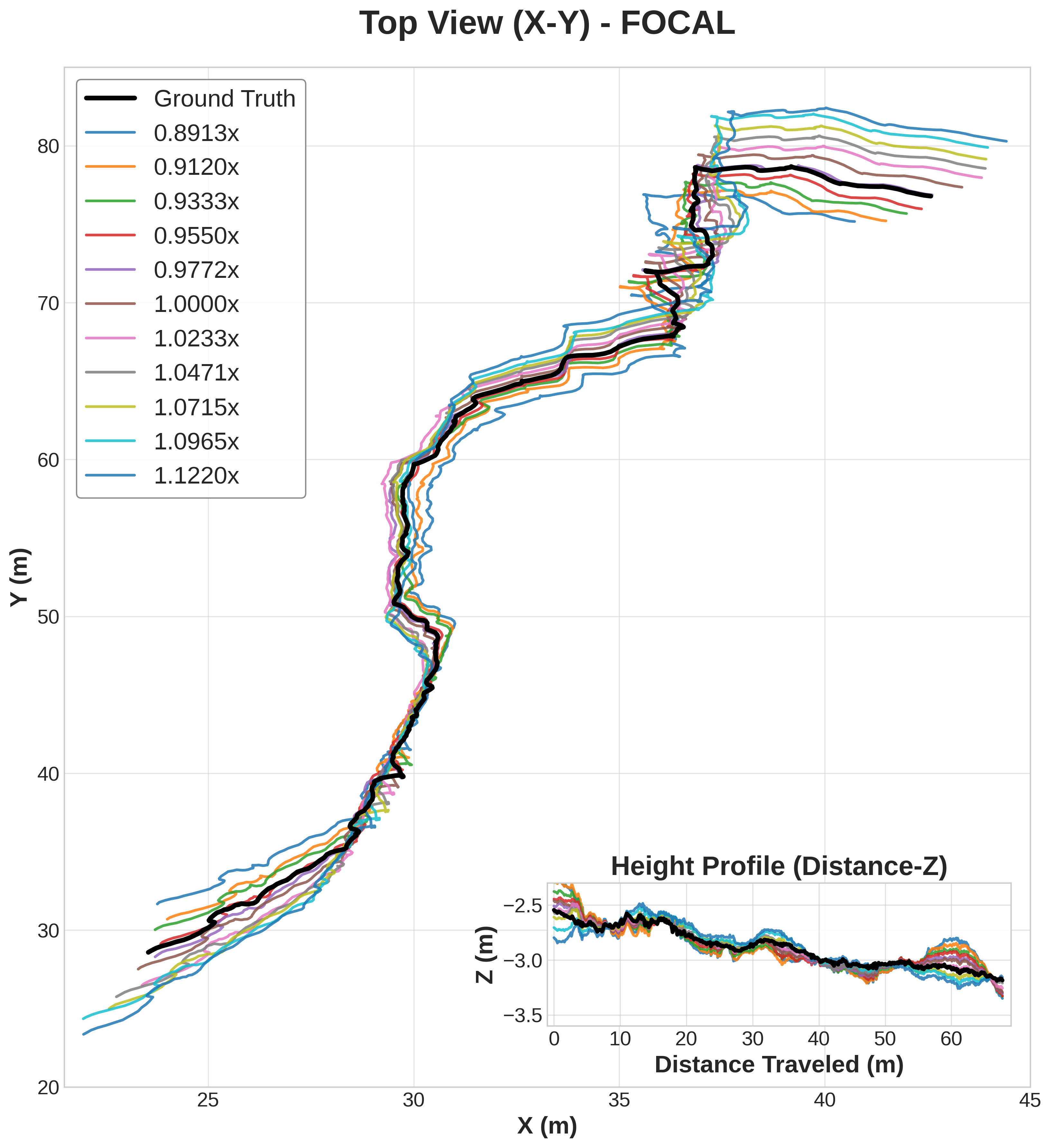}
    \caption{Focal Length ($f$)}
\end{subfigure}
\hfill
\begin{subfigure}{0.32\textwidth}
    \centering
    \includegraphics[width=\linewidth]{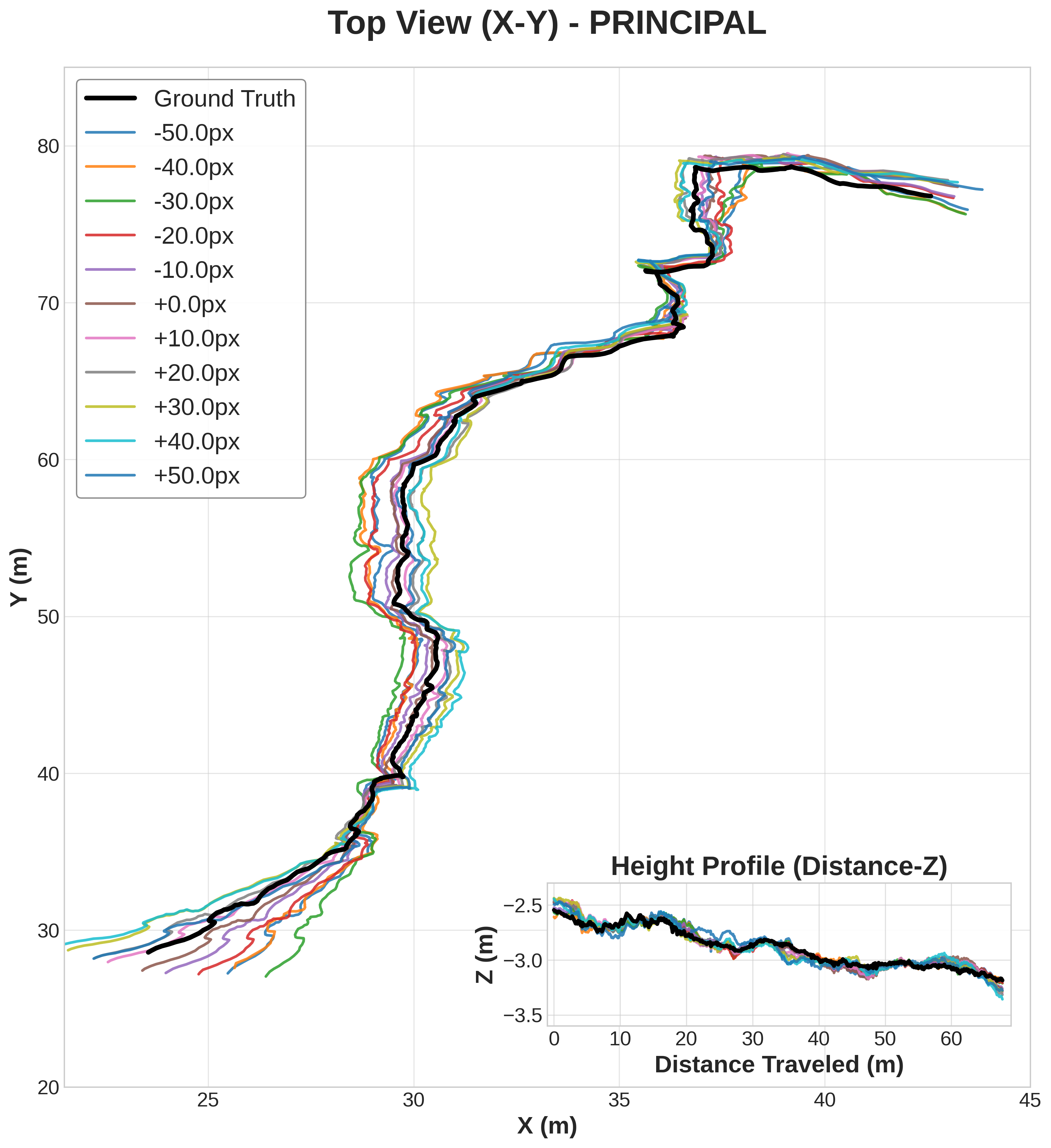}
    \caption{Principal Point ($c_p$)}
\end{subfigure}
\hfill
\begin{subfigure}{0.32\textwidth}
    \centering
    \includegraphics[width=\linewidth]{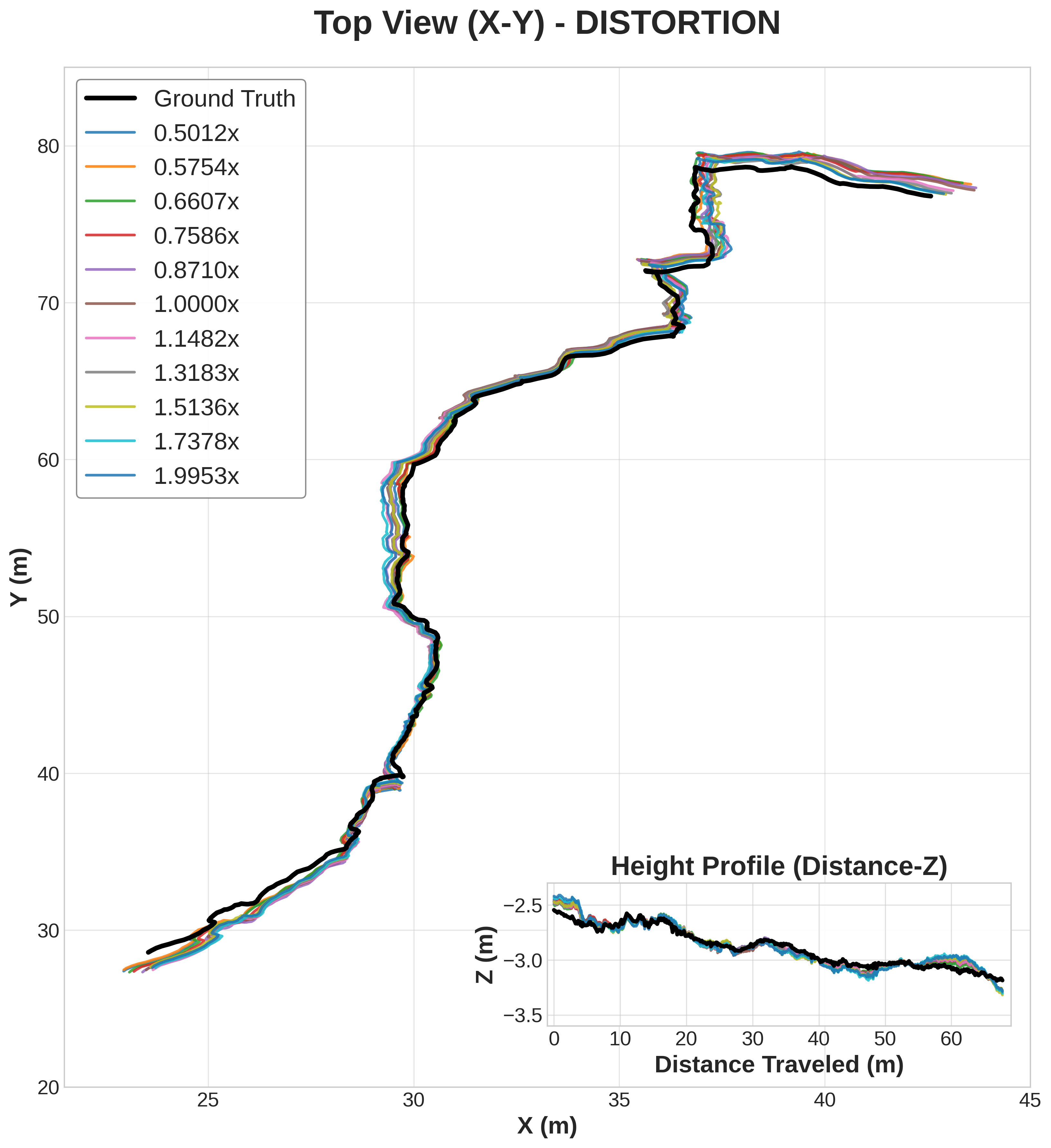}
    \caption{Distortion ($k_1$)}
\end{subfigure}

\caption{Gallery of trajectory deformations under nine distinct sensor perturbations of SchurVINS. While SchurVINS maintains high resilience against IMU stochastic noise and bias, it exhibits acute sensitivity to extrinsic miscalibration, where even minor misalignments in rotation ($R_{IC}$) precipitate rapid estimator divergence. Furthermore, inaccuracies in camera intrinsics—specifically the focal length ($f$) and principal point ($c_p$)—induce significant geometric residuals , leading directly to characteristic trajectory offsets and scale drift.}
\label{fig:nine_noise_trajectories}
\vspace{-10pt}
\end{figure*}

\section{DISCUSSION}

\subsection{Sensor Perturbation and Calibration Sensitivity Analysis}
The comprehensive sensitivity analysis delineated in Fig. \ref{fig:sensitivity_grid} elucidates fundamentally distinct failure mechanisms across the evaluated VIO frameworks, which are categorized herein into algorithm-specific behaviors, dimensional sensitivities, and operational deployment guidelines. To empirically characterize the impact of these perturbations, the qualitative deformations of trajectories exhibiting maximum estimator drift are visualized in Fig. \ref{fig:nine_noise_trajectories}, revealing that IMU accelerometer noise ($\sigma_a$) serves as a primary driver of unrectifiable quadratic position drift that exceeds the correction capabilities of visual update cycles.

\textbf{OpenVINS:} Despite demonstrating significant stability against visual parameter variances and extrinsic miscalibration, OpenVINS demonstrates acute vulnerability to inertial perturbations, particularly accelerometer bias ($b_a$) and noise ($\sigma_a$), where an abrupt estimator divergence occurs as these parameters exceed a $2.512 \times$ baseline threshold ($10^{0.4}$), corresponding to $b_a \approx 0.2512 \, m/s^2$ and $\sigma_a \approx 0.02512 \, m/s^2/\sqrt{\text{Hz}}$. As an EKF-based MSCKF architecture, its estimator consistency is strictly constrained by the noise covariance matrix $\mathbf{Q}$, such that innovation terms exceeding the first-order Taylor expansion's confidence interval trigger linearization failures. Unlike optimization-based paradigms, the EKF lacks a retrospective iterative refinement mechanism, such as Bundle Adjustment, to mitigate early-stage integration drift, resulting in unrecoverable state divergence as quadratic error accumulation ($p(t) = p_0 + v_0 t + \frac{1}{2}a_{\text{true}} t^2 + \frac{1}{2}b_a t^2$) overwhelms the visual measurement model.

\textbf{AirSLAM:} Conversely, AirSLAM maintains superior immunity to high-magnitude IMU noise through a learning-based front-end robust to motion-blur-induced degradations, yet its backend enforces rigid geometric constraints that amplify reprojection residuals linearly in the absence of robust cost functions. Furthermore, its instability in low-noise regimes suggests a susceptibility to numerical ill-conditioning when persistent excitation is insufficient to distinguish sensor bias from true kinematic states within the geometric state space.

\textbf{ORB-SLAM3:} This framework provides the most balanced state estimation by leveraging mature IMU pre-integration and Factor Graph Optimization (FGO) to effectively decouple inertial biases from kinematic states. Nevertheless, as a purely geometric formulation, it preserves a strict linear dependency on the intrinsic matrix $\mathbf{K}$, where offsets in focal length or the principal point induce systematic geometric residuals that cannot be suppressed through increased data redundancy.

\textbf{SchurVINS:} As shown in Fig.\ref{fig:nine_noise_trajectories} SchurVINS is resilient to IMU noise and exhibits high sensitivity to extrinsic, where minor misalignments trigger rapid estimator divergence due to the high sensitivity of its Schur-complement-based marginalization to the spatial-temporal alignment of gravity vectors and visual landmarks. Additionally, the fragility of its photometric consistency assumption renders the SVO-based front-end prone to tracking failure under the non-uniform illumination and mechanical vibrations characteristic of subterranean environments.

The cross-dimensional sensitivity analysis identifies fundamental architectural trade-offs in multi-sensor fusion: Factor Graph Optimization (FGO)-based frameworks demonstrate superior mitigation of inertial bias drift compared to EKF-based systems through iterative re-linearization, while OpenVINS and ORB-SLAM3 utilize robust cost functions to suppress the influence of spatial miscalibration, contrasted by the requirement of AirSLAM and SchurVINS for precise extrinsic rotation ($R_{IC}$) to preserve geometric consistency within the estimation backend. Moreover, while rigorous pre-processing ensures robustness against radial distortion ($k_1$), the MSCKF architecture of OpenVINS inherently provides geometric redundancy that desensitizes the estimator to absolute intrinsic precision, unlike the critical focal length and principal point dependencies observed in optimization-heavy frameworks.

To ensure optimal reliability in challenging environments, we propose the following guidelines:
\begin{enumerate}
    \item \textbf{Geometric Calibration:} For AirSLAM and SchurVINS, priority must be given to extrinsic ($R_{IC}$) and intrinsic ($c_p, f$) calibration accuracy to prevent excessive geometric residuals.
    \item \textbf{Inertial Noise Control:} For EKF-based systems like OpenVINS, the IMU noise floor and bias must be hardware-constrained within $2.5 \times$ of the nominal calibration. While for learned front-ends like AirSLAM, a minimum level of IMU excitation is beneficial to ensure numerical stability and prevent optimization ill-conditioning.
\end{enumerate}

\subsection{Dynamic and Occlusion Resilience Analysis}

The evaluation results illustrated in Fig. \ref{fig:occlusion_dynamics_heatmaps} demonstrate a systematic degradation in localization fidelity as a function of simultaneous increases in dynamics scale ($d$) and occlusion rate ($p$). This performance decay is primarily attributable to the severe attenuation of observable static landmarks and the concomitant increase in structured outlier interference during the estimator's measurement update phase.

\textbf{OpenVINS:} This framework exhibits the highest susceptibility to these perturbations, with a distinct failure envelope emerging at a 30\% occlusion rate where the Absolute Trajectory Error (ATE) RMSE peaks at 5.54~m, signifying catastrophic estimator divergence. This vulnerability originates from the Extended Kalman Filter (EKF) update mechanism; at this specific threshold, dynamic features associated with moving personnel coalesce into structured motion clusters that bypass standard geometric consistency checks. Consequently, these persistent outliers contaminate the state vector and precipitate rapid covariance inflation, ultimately leading to total tracking loss.

\textbf{AirSLAM:} This approach demonstrates enhanced survivability relative to EKF-based architectures, maintaining estimator stability at the 30\% occlusion threshold despite exhibiting non-negligible drift under extreme configurations, such as the 2.80~m RMSE recorded at $\ge$ 80\% occlusion and $d=20$. While learned feature descriptors afford a degree of resilience against dynamic feature clusters, the results suggest that a lightweight backend lacks the necessary redundancy to sustain high-precision localization when environmental static cues are significantly depleted.

\textbf{ORB-SLAM3:} This system maintains the most robust resilience profile, preserving an exceptionally low RMSE range of 0.04~m to 0.15~m even under the most demanding configurations of 90\% occlusion and $d=20$. This robustness is facilitated by a sophisticated backend employing Factor Graph Optimization (FGO) via Bundle Adjustment (BA) and multi-frame RANSAC filtering, which together enforce strict spatiotemporal geometric consistency to isolate dynamic outliers and prioritize peripheral static environmental structures for high-fidelity state estimation.

The comparative analysis of dynamic resilience highlights a fundamental disparity in how different estimation paradigms handle structured visual interference. Optimization-based architectures, by enforcing global geometric consistency over a sliding window, effectively marginalize dynamic outliers even when they occupy a significant portion of the field of view, whereas filter-based estimators exhibit a critical vulnerability threshold where dynamic feature clusters compromise the innovation term and lead to irreversible state divergence. This divergence is particularly pronounced in narrow subterranean corridors, where the spatial proximity of dynamic obstacles further intensifies the contamination of the visual measurement model, emphasizing the necessity for backend re-linearization to maintain estimator consistency under extreme dynamic stress.

To ensure optimal reliability in challenging environments, we propose the following guidelines:
\begin{enumerate}
    \item \textbf{Dynamic Sensor-Weighting:} EKF-based estimators should implement dynamic sensor-weighting strategies to prioritize IMU pre-integration as the system approaches the 30\% occlusion failure envelope to prevent state contamination.
    \item \textbf{Peripheral Feature Prioritization:} Feature tracking within narrow, occluded environments should prioritize peripheral image regions to maximize the utilization of stable environmental structures while naturally attenuating the impact of central dynamic disturbances.
\end{enumerate}

\section{CONCLUSION}

While ORB-SLAM3 achieves superior resilience through computationally intensive backend redundancy and Factor Graph Optimization (FGO), lightweight filtering-based paradigms like OpenVINS prioritize real-time efficiency within a narrower operational envelope sensitive to inertial noise, whereas learned architectures like AirSLAM effectively mitigate stochastic noise but remain susceptible to numerical ill-conditioning stemming from calibration inaccuracies. 

Estimator stability is fundamentally contingent upon the degradation source, as FGO-based re-linearization suppresses systematic internal residuals more effectively than the strict stochastic dependency of EKF architectures, while structured external outliers precipitate divergence by compromising the innovation term, highlighting the multi-modal nature of VIO robustness across sensor models and environmental constraints.

To ensure optimal reliability in subterranean deployments, we propose the following guidelines:
\begin{itemize}
    \item \textbf{Architectural Selection:} Prioritize FGO-based redundancy where computational resources allow; otherwise, ensure IMU noise floors remain hardware-constrained within a $2.5\times$ nominal boundary to prevent EKF linearization failure.
    \item \textbf{Calibration Hardening:} Enforce high-fidelity extrinsic ($R_{IC}, t_{IC}$) and intrinsic ($c_p, f$) calibration to minimize irreconcilable geometric residuals in optimization-heavy backends.
    \item \textbf{Dynamic Mitigation:} Employ dynamic sensor-weighting and peripheral feature tracking as occlusion approaches $30\%$ to attenuate central outlier interference and maximize reliance on inertial pre-integration.
\end{itemize}

\bibliographystyle{IEEEtran}
\bibliography{reference}

\end{document}